\documentclass[accepted]{uai2021}

\usepackage[american]{babel}

\usepackage{natbib} 
\usepackage{mathtools} 
\usepackage{booktabs} 
\usepackage{tikz} 

\usepackage{changepage}
\usepackage{algorithm} 
\usepackage{algpseudocode}
\usepackage{bbm}

\def\eqref#1{equation~\ref{#1}}
\def\Eqref#1{Equation~\ref{#1}}

\def\1{\bm{1}}

\def\eps{{\epsilon}}

\DeclareMathAlphabet{\mathsfit}{\encodingdefault}{\sfdefault}{m}{sl}
\SetMathAlphabet{\mathsfit}{bold}{\encodingdefault}{\sfdefault}{bx}{n}

\newcommand{\E}{\mathbb{E}}

\newcommand{\R}{\mathbb{R}}

\DeclarePairedDelimiterX{\infdivx}[2]{(}{)}{%
  #1\;\delimsize\|\;#2%
}

\DeclareMathOperator*{\argmax}{arg\,max}

\usepackage{xspace}
\usepackage[utf8]{inputenc}
\usepackage{amsthm}
\usepackage{amssymb}
\usepackage{mathtools}
\usepackage{hyperref}
\usepackage{appendix}
\usepackage[normalem]{ulem}

\usepackage{etoolbox}
\makeatletter
\patchcmd{\hyper@makecurrent}{%
    \ifx\Hy@param\Hy@chapterstring
        \let\Hy@param\Hy@chapapp
    \fi
}{%
    \iftoggle{inappendix}{
        \@checkappendixparam{chapter}%
        \@checkappendixparam{section}%
        \@checkappendixparam{subsection}%
        \@checkappendixparam{subsubsection}%
        \@checkappendixparam{paragraph}%
        \@checkappendixparam{subparagraph}%
    }{}%
}{}{\errmessage{failed to patch}}

\newcommand*{\@checkappendixparam}[1]{%
    \def\@checkappendixparamtmp{#1}%
    \ifx\Hy@param\@checkappendixparamtmp
        \let\Hy@param\Hy@appendixstring
    \fi
}
\makeatletter

\newtoggle{inappendix}
\togglefalse{inappendix}

\apptocmd{\appendix}{\toggletrue{inappendix}}{}{\errmessage{failed to patch}}
\apptocmd{\subappendices}{\toggletrue{inappendix}}{}{\errmessage{failed to patch}}

\newcommand\strike{\bgroup\markoverwith{\textcolor{red}{\rule[0.5ex]{2pt}{0.4pt}}}\ULon}

\newtheorem*{rep@theorem}{\rep@title}
\newcommand{\newreptheorem}[2]{%
\newenvironment{rep#1}[1]{%
 \def\rep@title{#2 \ref{##1}}%
 \begin{rep@theorem}}%
 {\end{rep@theorem}}}
\makeatother
\newreptheorem{theorem}{Theorem}
\newreptheorem{lemma}{Lemma}
\newreptheorem{assumption}{Assumption}

\newtheorem{corollary}{Corollary}[]
\newtheorem*{remark*}{Remark}

\newtheorem{definition}{Definition}

\usepackage{multirow, booktabs}
\usepackage{bm}
\usepackage{cancel}
\usepackage{color}
\newtheorem{prop}{Proposition}

\usepackage{natbib}
\usepackage{subfigure}
\usepackage{booktabs} 

\usepackage{times}

\title{Action Redundancy in Reinforcement Learning}

\author[*,1]{Nir Baram}
\author[*,1]{Guy Tennenholtz}
\author[1,2]{Shie Mannor}
\affil[*]{%
    \textbf{Equal Contribution}
}
\affil[1]{%
    Technion, Israel Institute of Technology
}
\affil[2]{%
    Nvidia Research
}
\begin{document}

\maketitle

\begin{abstract}
    Maximum Entropy (MaxEnt) reinforcement learning is a powerful learning paradigm which seeks to maximize return under entropy regularization. However, action entropy does not necessarily coincide with state entropy, e.g., when multiple actions produce the same transition.  Instead, we propose to maximize the transition entropy, i.e., the entropy of next states. We show that transition entropy can be described by two terms; namely, model-dependent transition entropy and \textbf{action redundancy}. Particularly, we explore the latter in both deterministic and stochastic settings and develop tractable approximation methods in a near model-free setup. We construct algorithms to minimize action redundancy and demonstrate their effectiveness on a synthetic environment with multiple redundant actions as well as contemporary benchmarks in Atari and Mujoco. Our results suggest that action redundancy is a fundamental problem in reinforcement learning. 
\end{abstract}

\section{Introduction}

\begin{figure}[t]
\centering
\includegraphics[width=0.3\textwidth, height=0.15\textheight, trim={0.cm 0.cm 0.cm 0.cm},clip]{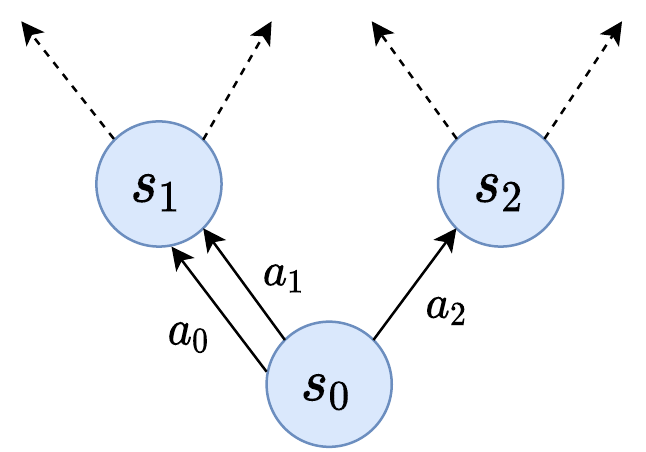}
\caption{\textbf{Suboptimality of Action Entropy}.
Plot depicts a deterministic MDP in which two actions transition to the same state. Maximizing the action entropy would bias transitions toward $s_1$, thereby producing suboptimal next-state entropy. Particularly, maximizing action entropy would result in a uniform distribution over the sets $\{a_0\}, \{a_1\}, \{a_2\}$, whereas maximizing transition entropy would result in a uniform distribution over $\{a_0, a_1\},\{a_2\}$.
}
\label{fig:regularization-illustration}
\end{figure}

Maximum Entropy (MaxEnt) algorithms are successful reinforcement learning (RL) approaches \citep{ziebart2008maximum, yin2002maximum, sac}, achieving superior performance through entropy regularization. The perturbed objective function promotes policies to maximize the reward while remaining stochastic, thereby encouraging exploration and stability. \citep{Ziebart2010Modeling}. Nevertheless, concurrent MaxEnt methods maximize action entropy, a proxy of the desired state entropy.
\par
Indeed, maximizing action entropy instead of state entropy may prove detrimental when these quantities diverge. Consider for example the case in which a subset of actions yields the same next state $s'$, as depicted in Figure~\ref{fig:regularization-illustration}. In this scenario, a uniform action distribution would increase the probability of visiting state $s'$ at the expense of not exploring other states.
\par
From an exploratory standpoint, generating uniform state visitation frequencies is an optimal strategy one can opt for in absence of any reward signal \citep{pitis2020maximum, pratissoli2020task,mutti2021task}. 
It turns out that finding a policy with a uniform state distribution can be cast as a convex optimization problem \citep{de2003linear}. However, this result holds only if the underyling MDP is fully specified. A less demanding setting is discussed in \citet{hazan2019provably}, where a provably efficient exploration algorithm requires access to a black-box planning oracle.
\par
In this work we propose to use transition entropy, i.e., next-state entropy, as a proxy for the state entropy. As the name implies, the next-state entropy measures the entropy a policy induces over states after a single step. Perhaps surprisingly, we show that estimating it is possible without having to learn a forward model for transitions.
\par
We begin by showing the expected transition entropy can be cast as the expectation of two terms, one representing model-dependent entropy, and the other \emph{action redundancy}. While the first term is attractive from a pure exploration point of view, it is undesirable when optimizing a controlled policy. For this reason, we focus on action redundancy and its implications. We derive and illustrate action redundancy in the stochastic and deterministic settings. We then derive value functions for the cumulative action-redundancy-to-go, allowing us to incorporate a novel regularization technique to both value-based \citep{watkins1992q} as well as actor-critic methods \citep{grondman2012survey}.

\section{Preliminaries}
\label{sec: preliminaries}

We assume a Markov Decision Process (MDP) ~\citep{puterman2014markov} which specifies the environment as a tuple ${\mathcal{M} = \langle \mathcal{S},\mathcal{A}, r, P, \gamma, \rho_0 \rangle}$, consisting of a state space $\mathcal{S}$, an action space $\mathcal{A}$, a reward function $r:\mathcal{S} \times \mathcal{A} \mapsto \R$, a transition probability function $P:\mathcal{S} \times \mathcal{A}\times \mathcal{S} \mapsto [0,1]$, a discount factor $\gamma \in (0,1)$,  and an initial state distribution $\rho_0: \mathcal{S} \mapsto [0,1]$. 

A policy $\pi: \mathcal{S} \times \mathcal{A} \mapsto [0, 1]$ interacts with the environment sequentially, starting from an initial state ${s_0 \sim \rho_0}$.  At time $t=0,1,\hdots$ the policy produces a probability distribution over the action set $\mathcal{A}$ from which an action $a_t \in \mathcal{A}$ is sampled and played. The environment then generates a scalar reward $r(s_t, a_t)$ and a next state $s_{t+1}$ is sampled from the transition probability function $P(\cdot | s_t, a_t)$. The value of a policy may be expressed by the state-action value function
${Q^\pi(s,a)=\mathbb{E}^\pi \big[ \sum\limits_{t=0}^\infty  \gamma^t r(s_t,a_t) | s_0=s, a_0=a \big]}$, 
or the state value function $v^\pi(s)=\mathbb{E}_{a \sim \pi}Q^\pi(s,a)$.

We use ${\rho_\pi(s,a)\triangleq (1-\gamma) \sum\limits_{t=0}^\infty \gamma^t P^\pi(s_t=s, a_t=a | s_0 \sim \rho_0)}$ and $\rho_\pi(s)\triangleq \sum_a\ \rho_\pi(s,a)$ to denote the discounted state-action and state visitation distributions of policy $\pi$, respectively. Finally we use $\mathbbm{1}_A(x)$ to denote the indicator function of a subset $A$ at $x$.

\textbf{Maximum Action Entropy.} The MaxEnt framework augments the expected return by introducing a discounted action entropy term
$$O_{AE}(s, \pi) = v^\pi(s) + \alpha \mathbb{H}(s, \pi),$$
where $\alpha \in \mathbb{R}_+$ and $\mathbb{H}(s,\pi)$ is the the discounted entropy, recursively defined by
\begin{align}\label{eq:discounted-action-entropy}
  \mathbb{H}(s,\pi)=\E_{a \sim \pi(\cdot | s)} \big[ g_\pi(a,s) + \gamma \E_{s' \sim P(\cdot | s,a)} \mathbb{H}(s',\pi)\big],
\end{align}
and $g_\pi(a,s) = -\log \pi(a|s)$. The discounted entropy can then be expressed recursively as \citep{nachum2017bridging}
\begin{align*}
O_{AE}(s,\pi) = \E_{a \sim \pi(\cdot | s)}\big[ &r(s,a) +
    \alpha g_\pi(a,s) + \\ &\gamma \E_{s' \sim P(\cdot | s,a)} O_{AE}(s',\pi) \big].
\end{align*}

\section{Transition Entropy}

We generalize the maximum action entropy framework to next-state transition entropy\footnote{Although state entropy is a preferred exploration criterion, it is by no means straightforward to estimate. As an intermediate step, we explore next-state transition entropy.}. Intuitively, actions resulting in similar transitions should be treated similarly \citep{tennenholtz2019natural,chandak2019learning}. Maximum action entropy translates to maximum transition entropy whenever no overlap between actions exists, i.e., whenever they induce distinct next states. In practice, this assumption is rarely met, as shown in Figure~\ref{fig:regularization-illustration}. It is thus reasonable to assume that maximum action entropy rarely coincides with maximum next-state transition entropy. 

In this section we formalize and define the transition entropy criterion, showing it can be cast as a sum of two terms~--~a model entropy term and an ``action redundancy" term.  Informally, redundancy in actions corresponds to their proximity w.r.t. their induced next-state transitions. In the sections that follow we provide methods to approximate action redundancy and propose to reweigh action scores of $O_{AE}$ (see Section~\ref{sec: preliminaries}) according to their redundancy, allowing for better exploration in terms of state entropy.

\begin{figure*}[t]
\centering
\includegraphics[width=0.97\textwidth]{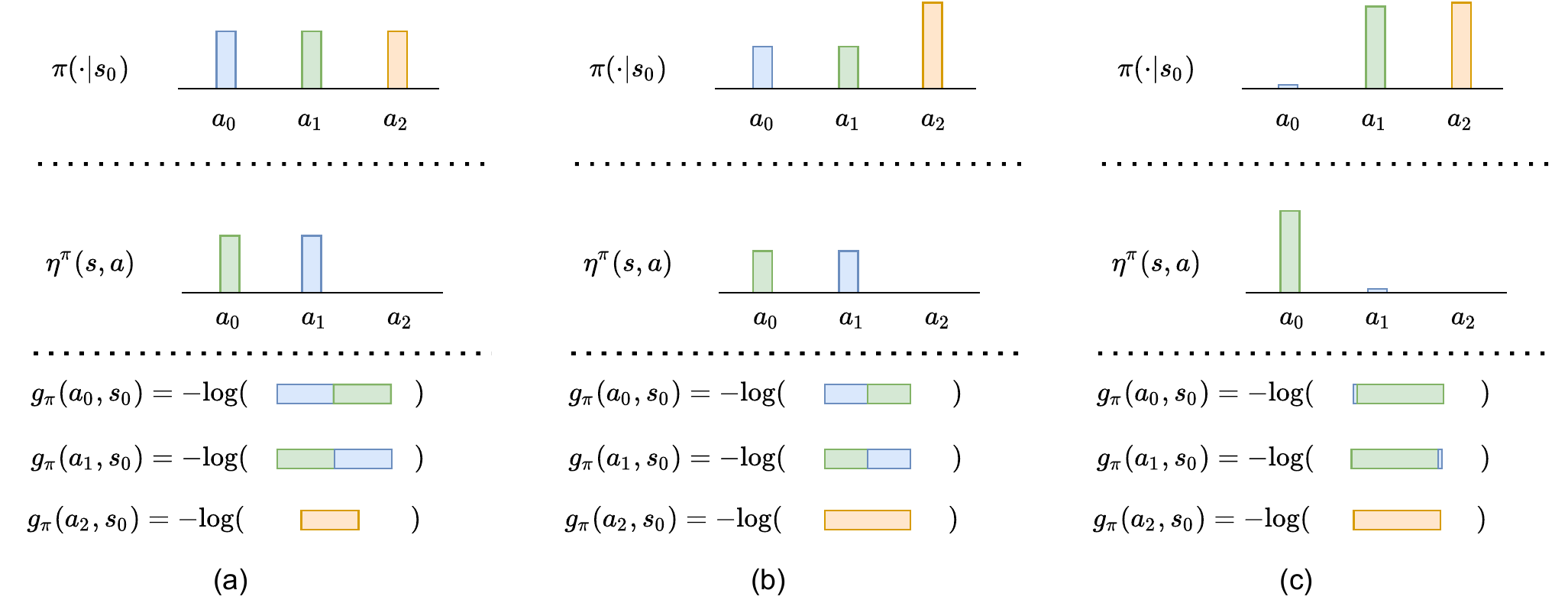}
\caption{\label{fig:acton_vs_nest_state}\textbf{Action Redundancy Score (ARS)}. 
Plot depicts ARS and $g_\pi$ w.r.t. the deterministic MDP in Figure~\ref{fig:regularization-illustration}. Here, both actions $a_0$ and $a_1$ transition to the same state. Top plot shows the action probabilities $\pi(a|s)$, middle plot the ARS, and bottom plot depicts the value of $g_\pi(s,a) = -\log(\pi(a|s) + \eta^\pi(s,a))$ (see Proposition~\ref{prop: determinstic g}). \textbf{(a)} The uniform policy which maximizes action entropy does not maximize transition entropy. \textbf{(b, c)} Examples of policies which maximize the transition entropy by minimizing overall action redundancy. }
\end{figure*} 

\subsection{From Action to Transition Entropy}
\label{sec:from-action-to-state-entropy}
We treat transition entropy as a myopic proxy for the discounted state entropy. As its name implies, transition entropy measures the entropy $\pi$ induces over states after a single step. As such, it is possible to estimate it from single-step transitions. To this end, we are interested in quantifying the discounted next-state transition entropy induced by a policy $\pi$. Let $P(s' | s, \pi) = \E_{a \sim \pi(s)} P(s'|s,a)$ denote the probability of transitioning from state $s$ to $s'$ following policy $\pi$. We define the transition entropy to-go as follows.
\begin{definition}
    We define the discounted transition entropy to-go following state $s$ by
    \begin{align*}
    \mathbb{F}(s,\pi)=
    -\sum_{t=0}^\infty \gamma^t  \E_{s_t \sim \rho_\pi}\E_{s' \sim P(\cdot | s_t, \pi)} \left[ \log P(s'|s_t, \pi) \Big| s_0=s \right].
  \end{align*}
\end{definition}
We can derive a recursive formula similar to \Eqref{eq:discounted-action-entropy} for the discounted transition entropy $\mathbb{F}(s,\pi)$, as shown by the following proposition.
\begin{prop}
\label{prop: f-recursive-form}
    Let $\mathbb{F}(s,\pi)$ as defined above. Then,
    \begin{align*}
        \mathbb{F}(s,\pi)= \E_{a \sim \pi(\cdot|s)} \left[ g_\pi(s,a) + \gamma  \E_{s' \sim P(\cdot | s,a)} \mathbb{F}(s', \pi) \right],
    \end{align*}
    where
    \begin{align}\label{eq:g-general}
        g_\pi(s,a) = - \E_{s' \sim P(\cdot | s, a)} \log P(s'|s,\pi).
    \end{align}
\end{prop}
Notice the resemblance of $\mathbb{F}(s,\pi)$ in Proposition~\ref{prop: f-recursive-form} to that of $\mathbb{H}(s, \pi)$ in \Eqref{eq:discounted-action-entropy} with a distinct definition for $g_\pi$. We can now define the Transition Entropy objective similarly to the Action Entropy objective in Section~\ref{sec: preliminaries} as
\begin{align*}
    O_{TE}(s,\pi) = v^\pi(s) + \alpha \mathbb{F}(s, \pi).
\end{align*}
In what follows, we decompose $g_\pi(s, a)$ in Equation~\ref{eq:g-general} to two terms: (1) a model-dependent entropy term, and (2) the action-redundancy term, for which we derive practical approximation methods for in both deterministic and stochastic MDPs.

\begin{table*}[t]
 
  \small\centering
  \begin{tabular}{lccr}

    \toprule

    {\bfseries Entropy Type} & {\bfseries Action Redundancy} & {\bfseries $\bm{g_\pi(s,a)}$}& {\bfseries Comments}\\
    \midrule\midrule[.1em]

    \multirow{1}{4cm}{Action Entropy} 
      & $0$
      & $- \log \pi(a|s)$ 
      & \citep{ziebart2008maximum}
      \\~\\
    \midrule[.1em] 
    
        
    \multirow{2}{4cm}{Transition Entropy \\ (Deterministic MDP)} 
      & $\eta^\pi(s,a)$
      & $- \log \big[\pi(a|s) + \eta^\pi(s,a) \big]$
      & $\eta^\pi(s,a)=\E_{\tilde{a} \sim \pi(\cdot | s)} \mathbbm{1}_{R_q(s, a)}(\tilde{a})$
      \\ & (ARS)& & (Proposition~\ref{prop: eta equivalence})\\
    \midrule[.1em] 

    \multirow{2}{4cm}{Transition Entropy \\ (Stochastic MDP)} 
      & $\zeta^\pi(s, a, s')$ 
      & $\mathcal{H}(s' | s,a) + \E_{s' \sim P(\cdot|s,a)} \zeta^\pi(s, a, s')$
      & $\zeta^\pi(s, a, s') = \log \frac{q^\pi(a|s,s')}{\pi(a|s)}$
      \\ & (ARR) & &
    \\
    \bottomrule
  \end{tabular}
  \caption{ \label{table:g-comparison} Action and transition entropy from an action redundancy perspective. ${\mathbb{F}(s,\pi)= \E_{a \sim \pi(\cdot|s)} \left[ g_\pi(s,a) + \gamma  \mathbb{F}(s', \pi) \right]}$.}
\end{table*}

\subsection{From Transition Entropy to Action Redundancy}

Recall the definition of $g_\pi$ in \Eqref{eq:g-general}, and notice that it can equivalently be written as follows
\begin{align}
    &g_\pi(s,a) = - \E_{s' \sim P(\cdot | s, a)} \log P(s'|s,\pi) \nonumber \\
    &
    =-\E_{s' \sim P(\cdot|s,a)} \left( \log P(s'|s,a)
    + \log \frac{P(s'|s, \pi)}{P(s'|s,a)} \right) \nonumber \\
    &=
    \underbrace{\mathcal{H}(s' | s,a)}_{\text{model entropy}}
    +
    \underbrace{D_{KL} (P(\cdot | s,a) || P(\cdot|s;\pi))}_{\text{action redundancy}}.
    \label{eq: g-decomposition}
\end{align}
We find that $g_\pi(s,a)$ is comprised of two meaningful terms. The first term represents the transition entropy given a state action pair $(s,a)$, whereas the second term represents the Kullback-Leibler divergence between the transition distribution given the state-action pair $(s, a)$ and the \emph{expected} next-state distribution induced by $\pi$.

The result in \Eqref{eq: g-decomposition} implies that two terms account for the \emph{transition score} of an action $a$ in a state $s$. First, the transition entropy of an action reflects the intrinsic (uncontrollable) entropy of the environment. Note that in deterministic or quasi-deterministic environments this term vanishes. Indeed, physics-based domains are mostly deterministic, allowing us to shift our focus to the second, controllable term. The second term reflects the ``importance" of an action w.r.t. the mean policy-induced transition. It is defined by the KL divergence between the state distribution an action produces and the expected distribution $\pi$ induces. As such, actions that achieve similar next-state distributions to the mean behavior would receive low scores. We thus recognize the latter as the action redundancy term.

While the intrinsic model entropy is appealing from an exploratory viewpoint, it is undesired from a control perspective \citep{pong2019skew}. In the remainder of the paper, we will focus on the second term, where we explain its meaning in terms of action redundancy and present tractable methods to approximate it.

\section{Action Redundancy}

In the previous section we showed that the expected next-state entropy to-go can be written as the sum of a model entropy term and an action redundancy term. While the former is intrinsic to the given environment, the latter is policy-dependent, and can thus be optimized to increase next-state entropy. In this section we derive expressions for action redundancy in both the deterministic and stochastic settings and design approximation methods. In the sections that follow we will construct algorithmic solutions based on these methods and demonstrate their effectiveness to eliminate redundancy in various domains.

\subsection{Deterministic Case}
\label{sec: deterministic case}
To better understand the implications of action redundancy and its relation to action entropy, we first turn to analyze the special case of deterministic MDPs, i.e., deterministic transitions. As we will see, in the deterministic case the model entropy vanishes, and action redundancy is reduced to an action equivalence set.

Assume a deterministic MDP with transition function ${f: \mathcal{S} \times \mathcal{A} \rightarrow \mathcal{S}}$. We begin by defining the action redundancy score (ARS) of an action $a \in \mathcal{A}$.
\begin{definition}
  Let $\pi$ be a policy. The action redundancy score (ARS) of action $a$ at state $s$ is defined by
  \begin{align}
  \label{eq: eta definition}
      \eta^\pi(s,a) 
      =
      \E_{\tilde{a} \sim \pi(\cdot | s)} \mathbbm{1}_{R(s, a)}(\tilde{a}),
  \end{align}
  where $R(s, a) = \left\{ \tilde{a} \in \mathcal{A}: \tilde{a} \neq a, f(s, \tilde{a}) = f(s,a) \right\}$.
\end{definition}
The ARS measures the total probability measure that $\pi$ assigns to other actions, $\tilde{a} \neq a$, such that $f(s,\tilde{a})=f(s,a)$. Indeed, whenever $\mathcal{A}$ is discrete we have that
\begin{align*}
    \eta^\pi(s,a)=\underset{\underset{f(s,\tilde{a})=f(s,a)}{\tilde{a} \neq a}}{\sum}\pi(\tilde{a}|s).
\end{align*}
The ARS is closely related to the action redundancy term in \Eqref{eq: g-decomposition}, as shown by the following proposition (see the appendix for proof).
\begin{prop}
\label{prop: determinstic g}
  Let $M$ be a deterministic MDP. Then
  \begin{align*}
      g_\pi(s,a) = - \log \left( \pi(a|s) + \eta^\pi(s,a) \right).
  \end{align*}
\end{prop}
Recall that for deterministic MDPs, $\mathcal{H}(s' | s,a) = 0$. Then, by Proposition~\ref{prop: determinstic g} and \Eqref{eq: g-decomposition} we have that 
\begin{align*}
    D_{KL} (P(\cdot | s,a) || P(\cdot|s;\pi)) = - \log \left( \pi(a|s) + \eta^\pi(s,a) \right).
\end{align*} 
This result suggests that by adding $\eta^\pi(s, a)$ to the traditional entropy bonus, $\log \pi(a|s)$, we essentially maximize the action entropy of a \textbf{modified action space} in which all redundant actions are unified, as depicted in Figure~\ref{fig:acton_vs_nest_state}. Specifically, notice that if $\eta^\pi(s, a)=0, \forall s,a$ (i.e., no two actions are assigned the same transition) the ARS is reduced to the discounted action entropy, as stated by the following corollary.
\begin{corollary}
Let $M$ be a deterministic MDP such that $f(s,a) \neq f(s,a'), \forall s,a \in \mathcal{S} \times \mathcal{A}$. Then
\begin{align*}
    \mathbb{F}(s, \pi) = \mathbb{H}(s, \pi) \quad , \quad \forall s \in S.
\end{align*}
\end{corollary}

\textbf{Estimating $\boldsymbol{\eta^\pi(s,a)}$.} Proposition~\ref{prop: determinstic g} provides us with the basic building block for minimizing action redundancy (and thus maximizing transition entropy) using the scores $g_\pi(s,a) = - \log \left( \pi(a|s) + \eta^\pi(s,a) \right)$. Nevertheless, estimating the ARS, $\eta^\pi(s,a)$, entails two major difficulties. First, as the transition function $f(s, a)$ is unknown, one must construct a forward model of $f(s,a)$. Second, a proper metric over states must be designed to execute the comparison over transitions. It turns out that we can sidestep these issues using an action equivalence relation.

Denote by $q^\pi(a | s, s')$ the posterior action distribution given that the state $s$ transitioned to $s'$. The posterior action distribution can be used to estimate the ARS, as shown by the following proposition (see the appendix for proof).
\begin{prop}
\label{prop: eta equivalence}
    Let $M$ be a deterministic MDP, and let $\pi$ be a policy satisfying $\pi(a | s) > 0$ for all $s, a \in \mathcal{S} \times \mathcal{A}$. Then, for any two actions $a, \tilde{a} \in \mathcal{A}$ and state $s \in \mathcal{S}$ we have
    \begin{align*}
        f(s,a)=f(s,\tilde{a}) \iff q^\pi(a|s,f(s,\tilde{a})) \neq 0.
    \end{align*}
    Moreover the ARS can be written as
    \begin{align}
    \label{eq: eta equivalence}
        \eta^\pi(s,a) 
        =
        \E_{\tilde{a} \sim \pi(\cdot | s)} \mathbbm{1}_{R_q(s, a)}(\tilde{a}),
   \end{align}
  where $R_q(s, a) = \left\{ \tilde{a} \in \mathcal{A}: \tilde{a} \neq a, q^\pi(\tilde{a}|s,f(s,a)) > 0 \right\}$.
\end{prop}
Note the difference in definition of the set $R_q$ in \Eqref{eq: eta equivalence} from $R$ in equation \Eqref{eq: eta definition}. For discrete actions, we can equivalently write \Eqref{eq: eta equivalence} as
\begin{align*}
    \eta^\pi(s,a) = \underset{\tilde{a} \neq a}{\sum} \pi(\tilde{a}|s) \mathbbm{1}[q^\pi(\tilde{a}|s,f(s,a)) > 0].
\end{align*}
We find that $\eta^\pi$ is in fact a function of the action posterior support. Given a transition $(s, a, s')$, approximating $q$, and in particular its support, is in general a much easier task than identifying $f(s,a)$ (whenever $\left| \mathcal{S} \right| \gg \left| \mathcal{A} \right|$).

Proposition~\ref{prop: eta equivalence} lets us estimate the ARS using the posterior action distribution $q^\pi$. We can efficiently learn the ARS using a backward model, sidestepping the need to learn a complex forward model of the environment. In fact, learning the exact posterior is not needed, but rather, only identifying its support is required to calculate the ARS. Our reliance on a learned model is thus minimized, allowing us to enjoy a near model-free setup.

\begin{algorithm}[t!]
	\caption{MinRed SAC}
	\textbf{Input:} Parametric models $Q_\psi, \pi_\theta, q_\phi$ 
	\begin{algorithmic}[1]
	    \State \textbf{Init: } Empty replay buffer $\mathcal{D}$
		\For {$i=0, 1, \hdots, T$}
		    \For {$t=0, 1, \hdots, K$}
        		\State Sample $a_t \sim \pi_\theta(\cdot|s_t)$.
        		\State Receive $r(s_t, a_t), s_{t+1} \sim P(\cdot|s_t,a_t)$.
        		\State Add $\{ s_t, a_t, r_t, s_{t+1}, \pi_\theta(s_t)\}$ to replay $\mathcal{D}$. 
    		\EndFor
    		\For {$n=0, 1, \hdots, N$}
    		    \State Sample $(s_i, a_i, r_i, s'_i, \pi_i) \sim D$.
        		\State Update $Q_\psi$ with $r_i + \alpha \frac{\pi_\theta}{\pi_i}\zeta_\phi^{\pi_i}(s_i, a_i, s'_i)$.
        		\State Update policy $\pi_\theta$ according to policy gradient.
        		\State Update posterior $q_\phi$ according to \Eqref{eq:phi-optimization}.
    		\EndFor
		\EndFor
	\end{algorithmic} \label{alg:minred}
\end{algorithm}

\begin{figure*}[t!]
\centering     
\subfigure[]{\label{fig:synth-maxent}\includegraphics[width=0.33\textwidth]{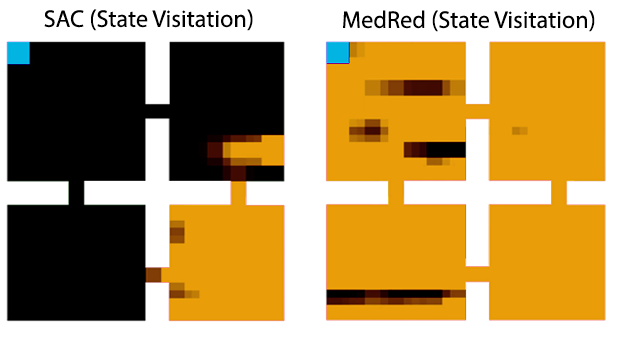}}
\subfigure[]{\label{fig:synth-hist}\includegraphics[width=0.33\textwidth]{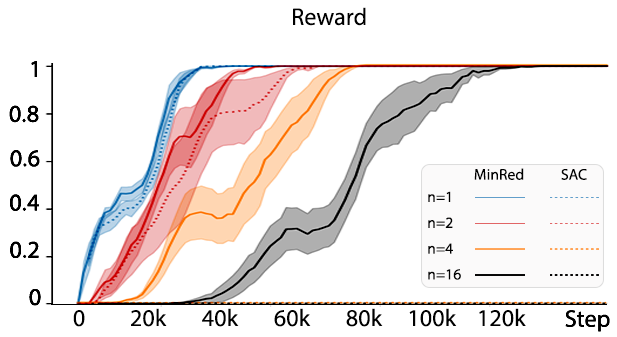}}
\subfigure[]{\label{fig:synth-reward}\includegraphics[width=0.33\textwidth]{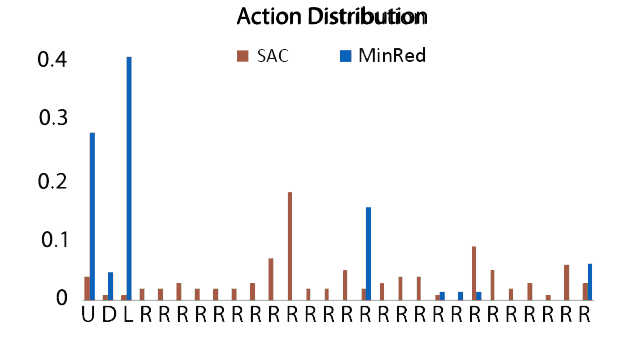}}

\caption{\label{fig:synthetic} \textbf{Synthetic Action Redundancy}. A four-room environment in which an agent is initialized at the bottom right corner and the goal is always located at the top left. The action space contains $n$ redundant ``Right" actions. \textbf{(a)} A heatmap of the visitation locations of maximum entropy (left) and minimum redundancy (right) agents. \textbf{(b)} Comparison of SAC (dotted) and MinRed (solid) agents for different values of $n$. As $n$ increases, the SAC agent fails to reach the goal. \textbf{(c)} Action selection histogram for $n=35$ shows that the policy practiced by the MinRed agent is far from uniform.}
\end{figure*}

\subsection{Stochastic Case}

In the general stochastic case, estimating $g_\pi$ amounts to estimating both the model entropy as well as the action redundancy terms. Estimating the model entropy in non-tabular settings is an open problem closely related to out-of-distribution detection and uncertainty estimation. As such, estimating it can amount to large errors unless carefully designed for the specific domain at hand. In contrast, the action redundancy term is dependent on $\pi$ at state $s$ and can be greedily and efficiently optimized. We therefore focus on action redundancy, noting estimation techniques for model entropy as an orthogonal direction for future work. 

Recall $q^\pi(a | s, s')$, the posterior action distribution defined in the previous subsection. Similar to the deterministic setting, we are interested in expressing the action redundancy term in \Eqref{eq: g-decomposition} using $q^\pi$, mitigating its estimation complexity. We define the action redundancy ratio (ARR) as follows.

\begin{definition}
    Let $\pi$ be a policy satisfying $\pi(a | s) > 0$ for all $s,a \in \mathcal{S} \times \mathcal{A}$. The action redundancy ratio (ARR) of action $a$ at state $s$ transitioning to state $s'$ is defined by
    \begin{align*}
        \zeta^\pi(s, a, s') = \log \frac{q^\pi(a|s,s')}{\pi(a|s)}
    \end{align*}
\end{definition}

\begin{algorithm}[t!]
	\caption{MinRed Q-Learning}
	\textbf{Input:} Parametric models $Q_\psi, q_\phi$, threshold $\delta > 0$
	\begin{algorithmic}[1]
	    \State \textbf{Init: } Empty replay buffer $\mathcal{D}$
		\For {$i=0, 1, \hdots, T$}
		    \For {$t=0, 1, \hdots, K$}
        		\State Select $a_t \in \arg\max_{a} Q_\psi(s_t, a)$ .
        		\State Receive $r_t$ and $s_{t+1} \sim P(\cdot|s_t,a_t)$ .
        		\For {$\bar{a} \in \mathcal{A}_\delta(s_t, s_{t+1}) \cup \{a_t\}$}
        		    \State Add $\{ s_t, \bar{a}, r_t, s_{t+1}\}$ to replay $\mathcal{D}$.
    		    \EndFor
    		\EndFor
    		\For {$n=0, 1, \hdots, N$}
        		\State Update $Q_\psi$ with reward $r(s,a)$.
        		\State Update posterior $q_\phi$ according to \Eqref{eq:phi-optimization}.
    		\EndFor
		\EndFor
	\end{algorithmic} \label{alg:minred qlearning}
\end{algorithm}

The ARR measures the overall dependence of action $a$ at state $s$ executed by $\pi$ and its corresponding next state $s'$. In fact, the ARR is defined by the pointwise mutual information\footnote{The pointwise mutual information of outcomes belonging to random variables $X, Y$ is defined by $\text{PMI}(x,y)=\log \frac{P(x,y)}{P(x)P(y)}$.} (PMI) of $a \sim \pi(\cdot | s)$ and $s' \sim P(\cdot | s, \pi)$ as 
\begin{align*}
    \zeta^\pi(s, a, s') 
    = 
    \text{PMI}(a, s' \mid s, \pi)
    =
    \log \frac{P(a,s'|s, \pi)}{\pi(a|s)P(s'|s, \pi)}.
\end{align*}
The ARR is also closely related to the ARS (see Section~\ref{sec: deterministic case}), as stated by the following corrolary.
\begin{corollary}
    Let $M$ be a deterministic MDP. Then
    \begin{align*}
        \zeta^\pi(s, a, s')
        =
        \begin{cases}
            -\log \left( \pi(a|s) + \eta^\pi(s,a) \right) &, s' = f(s,a) \\
            0 & ,\text{o.w.}
        \end{cases}
    \end{align*}
\end{corollary}
As such, the ARR generalized the ARS for stochastic MDPs. As before, the ARR requires estimating the action posterior distribution, $q^\pi$, a generally easier task than estimating $P(s' | s, a)$. The following proposition relates the ARS to $g_\pi$ (see the appendix for proof).
\begin{prop}
\label{prop: g estimation stochastic}
    Let $\pi$ be a policy satisfying $\pi(a | s) > 0$ for all $s,a \in \mathcal{S} \times \mathcal{A}$. Then
    \begin{align}
    \label{eq: g estimation stochastic}
       g_\pi(s,a) = 
        \mathcal{H}(s' | s,a)
        +
        \E_{s' \sim P(\cdot|s,a)} \zeta^\pi(s, a, s')
    \end{align}
\end{prop}
The above proposition lets us estimate $g_\pi$ using the expected ARS over next state transitions. While estimating the expectation in \Eqref{eq: g estimation stochastic} requires sampling multiple transitions from $s' \sim P(\cdot|s,a)$, we can utilize historical information to overcome this need (e.g., by sampling from a replay buffer). 

In the next section we propose to leverage an estimate of the action posterior $q^\pi$ for minimizing redundancy over actions in Actor-Critic and Q-learning variants. We then experimentally show in Section~\ref{sec:experiments} this approach is beneficial, and evidently crucial in domains in which action-redundancy is present.

\section{Algorithms for Minimizing Action Redundancy}
In the previous sections we defined transition entropy and showed its relation to action redundancy (see summary in Table~\ref{table:g-comparison}). We analyzed action redundancy in both deterministic as well as stochastic MDPs, based relations to the ARS and ARR, and proposed to estimate them by modeling the action posterior $q^\pi(a|s,s')$. This allows us to bypass the need to construct a forward model of the environment and derive efficient algorithms for minimizing action redundancy, as we show next.

\textbf{Estimation the action posterior.} We construct a parametric model for the posterior action distribution $q^\pi$. Specifically, given a set of past interactions with the environment, ${D=\{ s^{(i)},a^{(i)},s'^{(i)} \}_{i=1}^N}$, the maximum likelihood estimator, $\phi^*$, of $q_\phi(a|s,s')$ is given by:
\begin{align}\label{eq:phi-optimization}
\phi^* \in \argmax_\phi \sum_{i=1}^N \log q_\phi(a^{(i)}|s^{(i)},s'^{(i)})
\end{align}
We optimize \Eqref{eq:phi-optimization} by minimizing the cross entropy in a standard supervised-learning fashion.

\subsection{MinRed SAC}

We begin by deriving an off-policy algorithm which leverages action redundancy in a MaxEnt Actor-Critic framework. We propose MinRed SAC, shown in Algorithm~\ref{alg:minred}, which uses a replay buffer to store transitions and updates a critic w.r.t. an approximate ARR bonus. The ARR is estimated by our parametric model $q_\phi$ (\Eqref{eq:phi-optimization}) and updated with the critic. Alternatively, the ARR can be replaced by the ARS for deterministic domains. As the ARR depends on $\pi$, importance weights can be used to correct for biases in the updates. Specifically, the policy probabilities $\pi_i$ ared stored with the transitions in the replay buffer and correct for the difference using the importance ratio $\frac{\pi}{\pi_i}$.

\begin{figure*}[t!]
\centering     
\subfigure[DemonAttack]{\label{fig:macro-2}\includegraphics[width=0.23\textwidth]{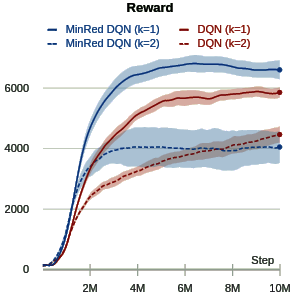}}
\subfigure[Breakout]{\label{fig:macro-1}\includegraphics[width=0.23\textwidth]{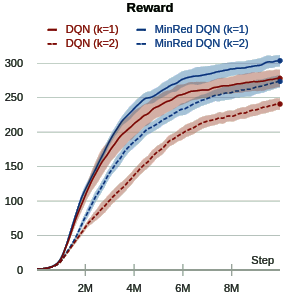}}
\subfigure[AirRaid]{\label{fig:macro-4}\includegraphics[width=0.23\textwidth]{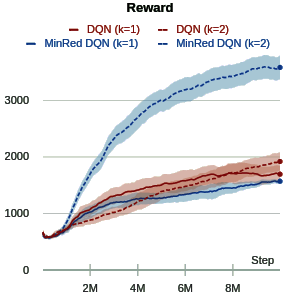}}
\subfigure[Seaquest]{\label{fig:macro-5}\includegraphics[width=0.23\textwidth]{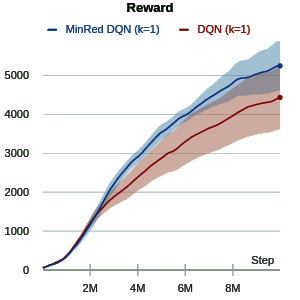}}

\caption{Results of MinRed DQN on Atari2600 benchmarks. We synthesized an action space consisting of all action sequences of length $k=2$ (i.e., cardinality $\left|\mathcal{A}\right|^k$). For Seaquest, the original action space (18 actions) was used, as we expect large action spaces to already contain redundancy. See Appendix~\ref{sec:implementation-details} for further details.}
\label{fig:atari results}
\end{figure*}

\subsection{MinRed Q-Learning}

A drawback of MinRed Actor Critic is its need of off-policy corrections for the shaped reward, as the ARR and ARS are updated by an old policy sampled from the replay buffer. To overcome this problem, we propose an alternative approach for minimizing action redundancy. Our approach, MinRed Q-Learning, shown in Algorithm~\ref{alg:minred qlearning}, uses an effective reduced action set to fight over-exploration of redundant actions. This method can be easily incorporated into any discrete algorithm (not necessarily MaxEnt).

Motivated by the dependence of the ARR and ARS on the action posterior $q^\pi$, we define the $\delta$-Redundant Action Set as follows. Given a state, next-state pair $(s,s')$ and a scalar $\delta > 0$, we define
\begin{align*}
    \mathcal{A}_\delta(s, s') = \left\{ a \in \mathcal{A} \, \mid \, q^\pi(a|s,s') > \delta \right\}.
\end{align*}
MinRed Q-Learning leverages the $\delta$-redundant action set $\mathcal{A}_\delta$ to accelerate exploration in the face of action redundancy. Specifically, whenever a new transition $(s, a, r, s')$ is sampled it is added to a replay buffer $\mathcal{D}$. Additionally, all equivalent actions (conditioned on the sensitivity threshold parameter~$\delta$), are also added to $\mathcal{D}$ i.e., ${\{(s, \bar{a}, r, s) \mid \bar{a} \in \mathcal{A}_\delta(s,s')\}}$. These synthetic transitions induce ``grouped exploration" of redundant actions, effectively reducing the dimensionality of the action space.

\section{Experiments}
\label{sec:experiments}

In this section we empirically evaluate our proposed algorithms on a series of domains. Specifically, we first construct a synthetic grid-like environment to analyze the effect of explicit action redundancy (i.e., duplication of actions) on performance. We then run experiments on standard discrete (Atari) and continuous control (Mujoco) benchmarks, showing contemporary algorithms suffer from the inherent redundancy.

\textbf{Synthetic Action Redundancy.} We wish to determine how well our method handles redundancy in a controlled synthetic environment. As such, we constructed a four-room environment in which the initial position of the agent and the goal were at two opposite corners (see Figure~\ref{fig:synthetic}). The agent earns a reward of $1$ at the goal and $0$ elsewhere. An episode ends whenever the agent achieves the goal or attempts $100$ steps. We constructed a synthetic redundant action space ${\mathcal{A} = \{ \texttt{Top, Left, Bottom,} \overbrace{\texttt{Right,\dots ,Right}} ^ {n}\}}$, for which the action \texttt{Right} was repeated $n$ times. To reach the goal on the far left and receive positive reward, the agent must counteract redundancy in actions pulling her to the right. 

We compared MinRed SAC and vanilla SAC for various values for $n$ (the number of repeats). Figure~\ref{fig:synthetic} depicts the reward of the agents as well as a visual representation of the areas visited by the agents. While SAC roams close to the origin, the MedRed agent reaches the goal, able to counteract action redundancy.

\textbf{Redundant Macro Actions.} Macro actions are commonly used to improve performance of RL agents. Macro actions are sequences of primitive actions, which, when chosen properly, may greatly shorten the effective horizon of the problem. However, designing macro-actions is a difficult task that is accomplished either through hierarchical reinforcement learning setups \citep{vezhnevets2017feudal}, or transfer learning algorithms \citep{konidaris2007building}. Unfortunately, using all action sequences of length $k$ results in an action space of cardinality $\left|\mathcal{A}\right|^k$. Such enlargement of the action space may be detrimental to the exploration process. Instead, we propose to leverage action redundancy to distinguish between the principal macro actions of the problem at hand.

We tested the MinRed Q-Learning agent on a series of games from the Atari2600 Arcade Learning Environment. We constructed a combinatorial action space on top of the original action space comprising of all action sequences (macro actions) of length $k$. While exploring such a large action space may be extremely challenging, through action redundancy the agent can potentially benefit from a short planning horizon at the cost of a moderately-sized action space. For environments with a small action space we extended the action space to action sequences of length $k=2$. For environments with a large amount of actions (e.g., Seaquest) we used the original action space, as redundancy was already evident.

Results, as presented in Figure~\ref{fig:atari results}, show improvement in performance when redundancy is accounted for. Significant improvement is evident when the number of actions increases due to our action sequence augmentation. Particularly, in various enviornments, when introducing action sequences, vanilla DQN did not see an increase in performance whereas our method showed either better or significantly superior performance. These results suggest that action redundancy is an intrinsic component of the reinforcement learning framework (in particular large action spaces), and overcoming it can greatly improve performance.
\begin{figure*}[t!]
\centering     
\subfigure[Halfcheetah (Dense)]{\label{fig:macro-2}\includegraphics[width=0.23\textwidth]{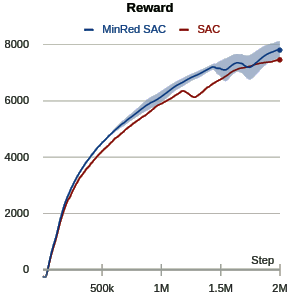}}
\subfigure[Halfcheetah (Sparse)]{\label{fig:macro-1}\includegraphics[width=0.23\textwidth]{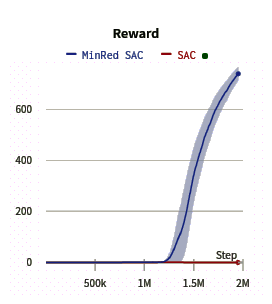}}
\subfigure[Humanoid (Dense)]{\label{fig:macro-4}\includegraphics[width=0.23\textwidth]{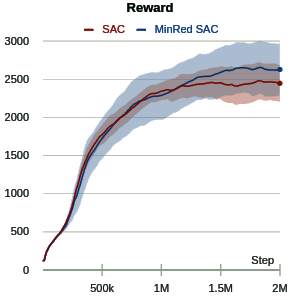}}
\subfigure[Humanoid (Sparse)]{\label{fig:macro-5}\includegraphics[width=0.23\textwidth]{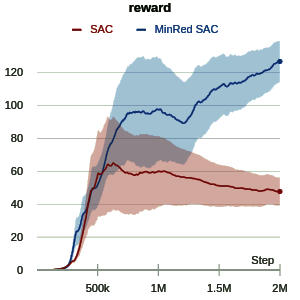}}

\caption{Results of MinRed SAC compared to SAC show an increase in performance in sparse reward domains, whereas dense reward domains remain mostly unaffected.}
\label{fig: mujoco results}
\end{figure*}

\textbf{Continuous Control.} We compared SAC against MinRed SAC on a series of continuous control benchmarks in Mujoco \citep{todorov2012mujoco}. Specifically, we constructed dense and sparse reward variants of the standard OpenAI-Gym Mujoco environments. In the standard, dense reward setting, the agent is rewarded at every time $t$ according to its absolute position, e.g., how much a humanoid has progressed in the $x$ direction. The agent is also penalized for taking actions with large magnitude. Conversely, in the sparse reward setting, the agent is only rewarded once it has traversed at least a \emph{threshold} distance, after which the original (dense) reward is given. 

In general, sparse reward settings are highly sensitive to efficient exploration. As such, we expect to observe significant improvement in the sparse reward variants. On the contrary, well designed dense rewards are less prone to exploration issues, but rather robust control. Indeed, results depicted in Figure~\ref{fig: mujoco results} show a significant improvement of MinRed SAC over SAC on sparse reward benchmarks, whereas performance is mostly unaffected in the dense setup. These results suggests that action redundancy is an inherent artifact of continuous control domains, obscured by concurrent benchmarks which utilize dense reward setups.

\section{Related Work}
\label{sec:related-work}
\textbf{Maximum Entropy.} Over the past decade MaxEnt algorithms have become a dominant solution for hard exploration environments \citep{ziebart2008maximum}. MaxEnt versions of popular RL algorithms were introduced \citep{haarnoja2017reinforcement}, often outperforming their baseline counterparts \citep{sac}. 

Orthogonal research has also focused on state-based exploration. Based on a planning oracle that finds the best policy under a well-specified reward signal, \citet{hazan2019provably} showed a provably efficient method that can achieve near-optimal state-entropy. On the other hand, maximizing a lower bound on the entropy of the steady-state distribution both removes the need for a planning oracle and allows learning fast mixing policies \citep{mutti2020intrinsically}. However, like other state-entropy methods that are oblivious to the geometry of states, this method is theoretically justified only for discrete states spaces. An adaptation to continuous domains requires using Geometric Entropy Maximisation (GEM) methods that respect the geometry of the Shannon entropy \citep{guo2021geometric} or alternative entropy measures \citep{mensch2019geometric}. 

Reaching diverse sets of states is also beneficial when learning` skills \citep{sutton1999between}. This is possible by using methods that promote diversity between trajectories \citep{lim2012autonomous, gajane2019autonomous, gregor2016variational}, or techniques that maximize the diversity within a single trajectory \citep{badia2020never}. The idea of maximum state entropy was also investigated in the context of multi-task RL \citep{zhang2017survey}. Training goal-dependent policies \citep{andrychowicz2017hindsight} to accomplish a diverse set of goals is possible by maximizing an information-theoretic surrogate function \citep{pong2019skew}. 

\textbf{Large Action Spaces.} Reasoning in environments with a large number of discrete actions or high dimensionality of continuous actions is a central obstacle for reinforcement learning. Various methods have been proposed to mitigate this difficulty, including action elimination \citep{even2003action,zahavy2018learn}, action embeddings \citep{dulac2012fast,tennenholtz2019natural,chandak2019learning}, and factorizations \citep{pazis2011generalized,dulac2012fast}. 

Our work provides an alternative perspective on the problem, suggesting that action redudnancy is present in problems of high dimensional action spaces and hard exploration. As such minimizing redundancy is a principal component for solving reinforcement learning tasks. Our approach reduces the action space effective dimensionality, thus allowing for more efficient learning. It can be readily applied to any reinforcement learning algorithm, as illustrated by our proposed MinRed methods (i.e., Algorithms~\ref{alg:minred}~and~\ref{alg:minred qlearning}).

\section{Conclusion}

This paper identifies and addresses the fundamental weakness of action redundancy in view of maximizing entropy. Through transition entropy, we isolated the problem and identified action redundancy as a key factor in increasing overall transition entropy. Our approach can be viewed as a MaxEnt framework in a modified action space that softly combines actions according to their induced transitions. We constructed two variants of contemporary algorithms to reduce action redundancy, showing signficantly improved performance on various benchmarks, and suggesting that action redundancy is indeed an inherent problem in RL. 

\begin{contributions}
    Nir Baram and Guy Tennenholtz contributed equally to this paper.
\end{contributions}

\begin{acknowledgements} 
This research was partially supported by the ISF under contract 2199/20.
\end{acknowledgements}

\bibliography{paper_bib}

\appendix
\onecolumn

\section{Missing Proofs}

We will use the following notation for our proofs.
\begin{align*}
    \mathcal{F}_\pi(s) = -\E_{s' \sim P(\cdot | s, \pi)} \log P(\tilde{s}|s, \pi).
\end{align*}
\textbf{Proof of Proposition~1.}

We have that
\begin{align*}
\mathbb{F}(s,\pi)&=
    -\sum_{t=0}^\infty \gamma^t  \E_{s_t \sim \rho_\pi}\E_{s' \sim P(\cdot | s_t, \pi)} \left[ \log P(s'|s_t, \pi) \Big| s_0=s \right]\\
  &=
  -\E_{s' \sim P(\cdot | s, \pi)} \log P(\tilde{s}|s, \pi)
  -\sum_{t=1}^\infty \gamma^t  \E_{s_t \sim \rho_\pi}\E_{s' \sim P(\cdot | s_t, \pi)} \left[ \log P(s'|s_t, \pi) \Big| s_0=s \right] \\
  &=
  \mathcal{F}_\pi(s) + \gamma \E_{s' \sim P(\cdot | s, \pi)}\mathbb{F}(s', \pi)
\end{align*}
Recall that $P(\cdot | s, \pi) = \E_{a \sim \pi(s)} P(s'|s,a)$. Then
\begin{align*}
    \E_{s' \sim P(\cdot | s, \pi)}\mathbb{F}(s', \pi)
    =
    \E_{a \sim \pi(s)} \E_{s' \sim P(\cdot | s, a)}\mathbb{F}(s', \pi).
\end{align*}
Similarly,
\begin{align*}
    \mathcal{F}_\pi(s) 
    = 
    -\E_{a \sim \pi(s)} \E_{s' \sim P(\cdot | s, a)} \log P(s'|s, \pi)
    =
    \E_{a \sim \pi(s)} g_\pi(s,a)
\end{align*}
where,
\begin{align*}
        g_\pi(s,a) = - \E_{s' \sim P(\cdot | s, a)} \log P(s'|s,\pi).
\end{align*}
    
This completes the proof.

\textbf{Proof of Proposition~2.}

We have that
\begin{align*}
        g_\pi(s,a) 
        &= 
        -\E_{s' \sim P(\cdot | s, a)} \log P(s'|s,\pi) \\
        &=
        - \mathbbm{1}_{s' = f(s,a)} \log P(s'|s,\pi)  \\
        &=
        - \mathbbm{1}_{s' = f(s,a)} \log \E_{\tilde{a} \sim \pi(s)} P(s'|s, \tilde{a}) \\
        &=
        - \log \E_{\tilde{a} \sim \pi(s)} P(s' = f(s,a)|s, \tilde{a}) \\
        &=
        - \log \left( \E_{\tilde{a}  \sim \pi(\cdot | s)} \left[ \pi(a | s) + \mathbbm{1}_{R(s, a)}(\tilde{a})\right] \right),
\end{align*}
where $R(s, a) = \left\{ \tilde{a} \in \mathcal{A}: \tilde{a} \neq a, f(s, \tilde{a}) = f(s,a) \right\}$. This completes the proof as $\eta^\pi(s,a) = \E_{\tilde{a} \sim \pi(\cdot | s)} \mathbbm{1}_{R(s, a)}(\tilde{a})$.

\textbf{Proof of Proposition~3.}

$f(s,a)=f(s,\tilde{a}) \iff q^\pi(a|s,f(s,\tilde{a})) \neq 0.$

Let $a, \tilde{a} \in \mathcal{A}$ and $s \in \mathcal{S}$. If $f(s,a)=f(s,\tilde{a})$ then trivially $q^\pi(a|s,f(s,\tilde{a})) > 0$. Conversely, if $q^\pi(a|s,f(s,\tilde{a})) \neq 0$ then using the fact that the transitions are deterministic and that $\pi_b(a|s) > 0$ for all $a \in \mathcal{A}$ we get that
\begin{align*}
    q^\pi(a|s,f(s,\tilde{a}))
    &=
    \frac{P^\pi(f(s,\tilde{a}), a | s) }{P^\pi(f(s,\tilde{a}) | s)} \\
    &=
    \frac{P(f(s,\tilde{a})| s, a) \pi(a | s) }{P^\pi(f(s,\tilde{a}) | s)}.
\end{align*}
By definition $P(f(s,a)| s, a) = 1$, therefore $q^\pi(a|s,f(s,\tilde{a})) \neq 0$ results in
\begin{align*}
    0 \neq P(f(s,\tilde{a})| s, a) \pi(a | s) = P(f(s,a)| s, a) \pi(a|s).
\end{align*}
Since $\pi(a|s) > 0$, the proof is complete.

\textbf{Proof of Proposition~4.}

Calculating action redundancy in stochastic MDPs amounts to measuring the KL divergence
\begin{align}\nonumber
    D_{KL} & \left( P(\cdot|s,a) \, || \, P(\cdot|s;\pi) \right) =
    \\\nonumber
    & \E_{s' \sim P(\cdot | s, a)} \log \bigg[ \frac{P(s'|s,a)}{\E_{\hat{a} \sim \pi(\cdot|s)}P(s'|s,\hat{a})} \bigg]
\end{align}
Applying Bayes Rule yields
\begin{align*}
    \E_{s' \sim P(\cdot | s, a)} \log \bigg[ \frac{P(s'|s,a)}{\E_{\hat{a} \sim \pi(\cdot|s)}P(s'|s,\hat{a})} \bigg]
    &=
    \E_{s' \sim P(\cdot | s, a)} \log \bigg[ 
    \frac{q^\pi(a|s,s')P^\pi(s'|s)}{\pi(a|s)\E_{\hat{a} \sim \pi(\cdot|s)}\frac{P^\pi(\hat{a}|s,s')P^\pi(s'|s)}{\pi(\hat{a}|s)}} \bigg] \\
    &=
    \E_{s' \sim P(\cdot | s, a)} \log \bigg[ 
    \frac{q^\pi(a|s,s')}{\pi(a|s)\E_{\hat{a} \sim \pi(\cdot|s)}\frac{P^\pi(\hat{a}|s,s')}{\pi(\hat{a}|s)}} \bigg] \\
    &=
    \E_{s' \sim P(\cdot | s, a)} \log \bigg[ 
    \frac{q^\pi(a|s,s')}{\pi(a|s)} \bigg],
\end{align*}
where in the last step we used the fact that
\begin{align*}
    \E_{\hat{a} \sim \pi(\cdot|s)}\frac{P^\pi(\hat{a}|s,s')}{\pi(\hat{a}|s)}
    =
    \int_{\hat{a}} \pi(\hat{a}|s) \frac{P^\pi(\hat{a}|s,s')}{\pi(\hat{a}|s)} d\hat{a}
    =
    \int_{\hat{a}} P^\pi(\hat{a}|s,s') d\hat{a}
    =
    1.
\end{align*}
This completes the proof.

\section{Implementation Details}
\label{sec:implementation-details}

\textbf{Specific Implementation Details.} A convolutional neural network is used to implement the policy and the critic. The layer specification of the network is provided in Appendix~\ref{sec:implementation-details}. We use an online actor critic algorithm as a baseline \citep{schulman2017proximal}. The baseline is modified to correct for action redundancy using Table~\ref{table:g-comparison}. We experiment with varying lengths of action sequences, $k$ and report the value used in the figure. 
\par
We run each task for 3 million steps and report an average of 5 seeds. We use the Adam optimizer \citep{kingma2014adam} with a learning rate of $0.0003$ and a batch size of 256 to train the algorithms. After collecting experience, the algorithm applies 4 consecutive stochastic gradient steps and clip the advantage as suggested in \citet{schulman2017proximal}. All other hyper-parameters are set to their default values as implemented in the Proximal-Policy-Optimization algorithm as part of the stable-baselines framework \citep{stable-baselines3}.

\subsection{MinRed Q-learning}
This section describes the implementation details and hyper-parameters used to train the MinRed DQN agent as well as other results.

\textbf{Network Architecture.}
The input to the policy is a concatenation of the last four frames, each of which is of the size $84 \times 84$. We implement the Q function using a neural network with the following specification.
\begin{table*}[h]
 
  \small\centering
  \begin{tabular}{lcc}

    \toprule

    {\bfseries Layer} & {\bfseries Size}  & {\bfseries Comments}\\
    \midrule\midrule[.1em]

    Conv2D & $4 \times 32 \times 8 \times 8$ & input channels $\times$ ouput channels $\times$ kernel size $\times$ kernel size \\
    \midrule[.1em] 

    ReLU & -- & \\
        \midrule[.1em] 
    
    Conv2D & $32 \times 64 \times 4 \times 4$ & \\
    \midrule[.1em] 
    
    ReLU & -- & \\
        \midrule[.1em] 
        
    Conv2D & $64 \times 64 \times 3 \times 3$ & \\
    \midrule[.1em] 
    
    ReLU & -- & \\
        \midrule[.1em] 
        
    Flatten & -- & \\
        \midrule[.1em] 
        
    Linear & $64$ & \\
    \midrule[.1em] 
    
    ReLU & -- & \\
        \midrule[.1em] 
        
    Linear & $64$ & \\
    \midrule[.1em] 
    
    ReLU & -- & \\
        \midrule[.1em] 
        
    Linear & $|A|$ & \\
    \midrule[.1em] 
    \bottomrule
  \end{tabular}
  \caption{MinRed DQN agent. \label{table:MinRed DQN-spec} A detailed description of the neural network function approximation architecture.}
\end{table*}

\textbf{Hyper-Parameters.}
The hyperparameters used to train the MinRed DQN agents are presented in Table~\ref{table:MinRed DQN-hyper-params}.

\begin{table*}[h]
 
  \small\centering
  \begin{tabular}{lcc}

    \toprule

    {\bfseries Name} & {\bfseries Value}  & {\bfseries Comments}\\
    \midrule\midrule[.1em]

    Batch size & $32$ &  \\
    \midrule[.1em] 

    Learning rate & $0.0001$ &  \\
    \midrule[.1em] 
    
    Buffer size & $150,000$ &  \\
    \midrule[.1em] 
    
    Total timesteps & $10,000,000$ &  \\
    \midrule[.1em] 
    
    Exploration initial value ($\eps$-greedy) & $1.$ &  \\
    \midrule[.1em] 
    
    Exploration final value ($\eps$-greedy) & $0.05$ &  \\
    \midrule[.1em]
    
    Learning starts & $50,000$ & iteration to start learning \\
    \midrule[.1em] 
    
    Regularization starts & $150,000$ & iteration to start loading synthetic transitions \\
    \midrule[.1em] 
    
    $\gamma$ & $0.99$ & Discount factor \\
    \midrule[.1em] 
    
    $\tau$ & $1$ & target network Polyak averaging\\
    \midrule[.1em] 
    
    Target network update interval & $1,000$ & \\
    \midrule[.1em] 
    
    \bottomrule
  \end{tabular}
  \caption{Hyper-parameters used to train the MinRed DQN agent. \label{table:MinRed DQN-hyper-params} A detailed description of the hyper-parameters used to train the MinRed DQN agents.}
\end{table*}

\textbf{Action Space Size.}
Table~\ref{table:MinRed DQN-action-space} describes the action space size used to train the MinRed DQN agents.
\begin{table*}[h]
 
  \small\centering
  \begin{tabular}{lcc}

    \toprule

    {\bfseries Environment} & {\bfseries $|A|$ (original)}  & {\bfseries $|\tilde{A}|$ (extended)}\\
    \midrule\midrule[.1em]

    Breakout & $4$ & $16$ \\
    \midrule[.1em] 
    
    DemonAttack & $6$ & $36$ \\
    \midrule[.1em]
    
    AirRaid & $6$ & $36$ \\
    \midrule[.1em]
    
    NameThisGame & $6$ & $36$ \\
    \midrule[.1em]
    
    \bottomrule
  \end{tabular}
  \caption{Action Space Size. \label{table:MinRed DQN-action-space} A detailed description of the action space size used by the MinRed DQN agents.}
\end{table*}

\begin{figure*}[h]
\centering     
\subfigure[DemonAttack]{\label{fig:mask-1}\includegraphics[width=0.24\textwidth]{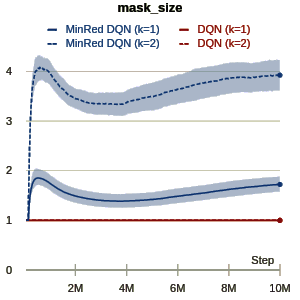}}
\subfigure[Breakout]{\label{fig:mask-2}\includegraphics[width=0.24\textwidth]{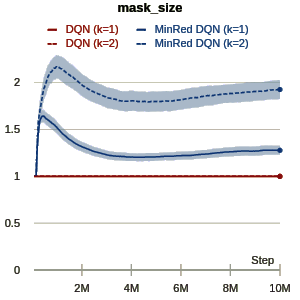}}
\subfigure[AirRaid]{\label{fig:mask-3}\includegraphics[width=0.24\textwidth]{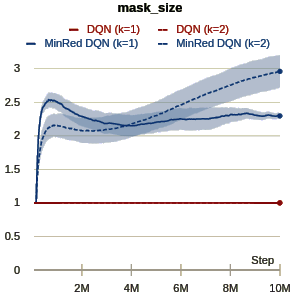}}
\subfigure[Seaquest]{\label{fig:mask-4}\includegraphics[width=0.24\textwidth]{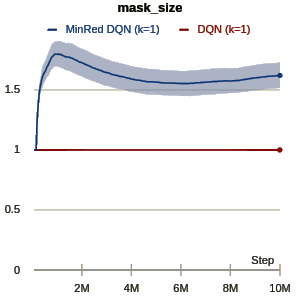}}

\caption{\textbf{Redundancy Size.} The graph shows the average number of synthetic transitions created per single transition sampled from a real environment.}
\label{fig:mask-size}
\end{figure*}

\textbf{Redundancy Size.}
Figure~\ref{fig:mask-size} shows the average mask size of the MinRed DQN agent. That is, the number of synthetic transitions loaded to the replay buffer. Note that the mask-size of the baseline method is identically 1.

\newpage
\subsection{Minimum Redundancy Soft Actor Critic}

This section describes the implementation details and hyper-parameters used to train the minimum redundancy agent as well as other results.

\textbf{Network Architecture.}
The input to the policy is a vector indicating the positions and velocities of the joints of the agent. Using a neural network with the following specification, we implement both the soft Q function and the policy. The input to the Q-function is a concatenation of $s$ and $a$ and the output is a single value corresponding to $Q(s,a)$. The policy network accepts a state $s$ as input and produces two vectors of size $|A|$. The two outputs represent the first two moments of a Gaussian distribution, from which an action is sampled. The action in this case, represents the forces to apply to each joint. The action is squashed to the range of $[-1,1]$ using a $\texttt{Tanh}$ layer. Table~\ref{table:minred-sac-spec} lists the parametric layers of the neural network function approximation.

\begin{table*}[h!]
 
  \small\centering
  \begin{tabular}{lcc}

    \toprule

    {\bfseries Layer} & {\bfseries Size}  & {\bfseries Comments}\\
    \midrule\midrule[.1em]

    Linear & $256$ &  \\
    \midrule[.1em] 

    ReLU & -- & \\
        \midrule[.1em] 
    
    Linear & $256$ &  \\
    \midrule[.1em] 
    
    ReLU & -- & \\
        \midrule[.1em] 
        
    Linear & $|A|$ &  action space size\\
    \midrule[.1em] 
    
    \bottomrule
  \end{tabular}
  \caption{MinRed DQN agent. \label{table:minred-sac-spec} A detailed description of the neural network function approximation architecture.}
\end{table*}

\textbf{Hyper-Parameters.}
The hyperparameters used by the MinRed agents are shown in Table~\ref{table:minred-hyper-params}.

\begin{table*}[h!]
 
  \small\centering
  \begin{tabular}{lcc}

    \toprule

    {\bfseries Name} & {\bfseries Value}  & {\bfseries Comments}\\
    \midrule\midrule[.1em]

    Batch size & $256$ &  \\
    \midrule[.1em] 

    Learning rate & $0.0003$ &  \\
    \midrule[.1em] 
    
    Buffer size & $100,000$ &  \\
    \midrule[.1em] 
    
    Total timesteps & $2,000,000$ &  \\
    \midrule[.1em] 
    
    Learning starts & $10,000$ & iteration to start learning \\
    \midrule[.1em] 
    
    train frequency & $1000$ & Update the model every train-freq steps \\
    \midrule[.1em] 
    
    gradient-steps & $1000$ & How many gradient steps to do after each rollout \\
    \midrule[.1em] 
    
    $\gamma$ & $0.99$ & Discount factor \\
    \midrule[.1em] 
    
    $\tau$ & $0.05$ & target network Polyak averaging\\
    \midrule[.1em] 
    
    Target network update interval & $1$ & \\
    \midrule[.1em] 
    
    \bottomrule
  \end{tabular}
  \caption{Hyper-parameters used to train the Minimum Redundancy agents. \label{table:minred-hyper-params} A detailed description of the hyper-parameters used to train the minimum redundancy agents.}
\end{table*}

\textbf{Entropy Coefficients.}
Table~\ref{table:minred-ent-coefs} describes the entropy and action redundancy coefficients $\alpha$, and $\lambda$ respectively used for different environments. As a thumb rule, we set $\lambda=0.01\alpha$.

\begin{table*}[h!]
 
  \small\centering
  \begin{tabular}{lccc}

    \toprule

    {\bfseries Environment} & {\bfseries $\alpha$ (entropy coefficient)}  & {\bfseries $\lambda$ (action redundancy)} & {\bfseries sparsity threshold} \\
    \midrule\midrule[.1em]

    Ant-v2 & $0.1$ & $0.001$ & -- \\
    \midrule[.1em] 
    
    HalfCheetah-v2 & $0.05$ & $0.05$ & -- \\
    \midrule[.1em] 
    
    Humanoid-v2 & $0.1$ & $0.001$ & -- \\
    \midrule[.1em] 
    
    HumanoidStandUp-v2 & $1.0$ & $0.01$ & -- \\
    \midrule[.1em] 
    
    Walker-v2 & $0.1$ & $0.1$ & -- \\
    \midrule[.1em] 
    
    SparseAnt-v2 & $0.05$ & $0.005$ & 15 \\
    \midrule[.1em] 
    
    SparseHalfCheetah-v2 & $0.05$ & $0.005$ & 19.5 \\
    \midrule[.1em] 
    
    SparseHumanoid-v2 & $0.05$ & $0.0005$ & 0.6 \\
    \midrule[.1em] 
    
    SparseHumanoidStandUp-v2 & $0.05$ & $0.005$ & -- \\
    \midrule[.1em] 
    
    SparseWalker-v2 & $0.1$ & $0.001$ & 15 \\
    \midrule[.1em] 
    
    \bottomrule
  \end{tabular}
  \caption{Entropy Coefficients. \label{table:minred-ent-coefs} The entropy coefficient $\alpha$ is set by running the baseline method, soft-actor-critic, with a tenable entropy coefficient and taking the average value of the last 100k iterations. The action redundancy coefficient, $\lambda$ is taken to be $\lambda = 0.01\alpha$ unless otherwise mentioned.}
\end{table*}

\end{document}